# Experience-based Optimal Motion Planning Algorithm for Solving Difficult Planning Problems Using a Limited Dataset*

Ryota Takamido and Jun Ota

*Abstract*— This study aims to address the key challenge of obtaining a high-quality solution path within a short calculation time by generalizing a limited dataset. In the informed experience-driven random trees connect star (IERTC*) process, the algorithm flexibly explores the search trees by morphing the micro paths generated from a single experience while reducing the path cost by introducing a re-wiring process and an informed sampling process. The core idea of this algorithm is to apply different strategies depending on the complexity of the local environment; for example, it adopts a more complex curved trajectory if obstacles are densely arranged near the search tree, and it adopts a simpler straight line if the local environment is sparse. The results of experiments using a general motion benchmark test revealed that IERTC* significantly improved the planning success rate in difficult problems in the cluttered environment (an average improvement of 49.3% compared to the state-of-the-art algorithm) while also significantly reducing the solution cost (a reduction of 56.3%) when using one hundred experiences. Furthermore, the results demonstrated outstanding planning performance even when only one experience was available (a 43.8% improvement in success rate and a 57.8% reduction in solution cost).

## I. Introduction

Sampling-based motion planning has achieved significant success in robotic motion planning [1]. It attempts to sample a valid state in a configuration space (C-space) and to find a feasible path where the robot can move from the initial to the goal configuration while avoiding obstacles in the environment. The advantage of a sampling-based planner is its ability to address high-dimensional planning problems without creating strict environmental models.

To efficiently use motion planning algorithms in practical situations, the key challenge is to balance the solution quality and calculation time. While the "non-optimal" motion planning algorithms, such as the rapidly exploring random trees connect (RRTC) [2], can find a solution path quickly, the quality of the solution tends to be worse. However, optimal motion-planning algorithms, such as RRT* [3], often struggle with the explosion of calculation time in a cluttered environment with many surrounding obstacles. Therefore, previous studies have focused on obtaining a better solution for difficult planning problems within a shorter calculation time [4].

To address this dilemma, one major approach is to use information from past planning results (i.e., experiences) to quickly find a high-quality solution. Because a robot often moves in a similar but slightly different environment in a practical situation, the past solution path can be an efficient clue for solving a new problem. Therefore, many previous studies adopt the "learning from experiences" strategy to effectively reuse past experiences in new problems, using various techniques such as statistical modeling [5], reinforcement learning [6], or deep learning [7].

However, the issue with the above approach is the difficulty in generalizing a limited number of experiences in different environments [4]. Most of the current learning-based approaches require large datasets to handle complex and varied planning problems. However, this creates another dilemma: developing an algorithm to solve difficult planning problems requires extensive experience in solving those problems, as "from-scratch" motion planners often struggle with them. From a practical perspective, generalizing a limited number of experiences is ideal to solve numerous difficult problems.

Therefore, as described in this section, the key challenge of the current motion planning algorithm is to find a high-quality solution within a short calculation time using a limited dataset.

## II. Related Work

### A. Optimal Motion Planning

In optimal motion planning, most existing algorithms are derived from the RRT* algorithm [3]. RRT* reduces the solution cost, which is typically represented by the length of the solution path in C-space, by re-wiring the connection between the nodes in the search tree for every sampling. Because the original RRT* struggles at a slow convergent speed, the main target of its derivative algorithms is to increase the convergent speed while maintaining solution quality. For example, RRT*-Smart [8] and Quick-RRT* [9] improved the RRT* algorithm by modifying the re-wiring process to increase the convergence speed, and other algorithms adopted a lazy collision checking strategy to reduce the calculation cost [10, 11].

The Informed RRT* [12] is another milestone in this field of research. In this algorithm, an RRT*search was first conducted until an initial solution was obtained. Subsequently, after updating the best cost by finding a new solution, it prunes unnecessary nodes in the search tree and restricts the search region in the C-space using cost heuristics. These improvements enable the planner to intensively explore the search tree in the informative region in the C-space and quickly reduce the solution cost [12]. The concept of informed sampling has been widely adopted in recent studies. For example, Gammell et al. [13] proposed batch-informed tree star (BIT*), which effectively integrates tree-based and graph-based search algorithms by adopting an informed sampling strategy. Strub and Gammell [14] added further improvements to the BIT* algorithm and proposed an adaptive information tree star (AIT*) that uses an asymmetric bidirectional search to simultaneously estimate and exploit problem-specific cost heuristics. There are also many optimal motion planning algorithms, as reviewed in [4]; however, many of these algorithms have focused on how to effectively change the connections in the search tree (RRT*-based) or how to effectively restrict the sampling region in the C-space (Informed RRT*-based).

### B. Experience-based Motion Planning

While the algorithms reviewed in the previous section adopt a planning-from-scratch strategy, which does not use any prior information, the use of previous experiences further improves the performance of those planners.

There are two main approaches in this research direction. One is a learning-based approach that builds a statistical model or trains a machine-learning algorithm based on previous experiences and integrates them into the planning process [15]. For example, Wang et al. [16] proposed a Gaussian mixture regression (RRT*) that performs informed sampling based on a GMR model built on previous experiences. They also developed a Neural RRT* [17] that introduced a convolutional neural network into the RRT* sampling process to effectively process environmental image information. Hung et al. [18] recently proposed a Neural Informed RRT* that generates guide sampling into a restricted region for informed sampling using a neural network technique. Other studies have also introduced machine learning techniques into the optimal motion planning process to increase the convergence speed of RRT*-based sampling [19, 20].

However, the problem with learning-based methods is the difficulty in generalizing a limited number of experiences to various environments. As reviewed in [4], most learning-based algorithms are restricted to environments similar to those of the training data, thereby performing poorly when transferred to more complex conditions or those with larger differences.

Conversely, the other approach is to use the "retrieve and repair" strategy [21-23]. In contrast to learning-based approaches, it does not require explicitly building a statistical model or pretraining the machine learning algorithm with a large dataset. It extracts a single experience path that is planned in the most similar problem to the current one and deforms it to fit the new environment. For example, in the Lightning framework proposed in [21], the most related experience was selected from the path library based on the similarity of the initial and goal configurations and adjusted into new problems using the RRTC-based sampling process. In this research direction, the experience-driven random tree (ERT) and its bidirectional version ERTC (ERT-Connect) [23] have the highest ability to generalize a limited number of experiences to various environments. These algorithms first map the experience into a new environment and then divide it into several micro-experiences. Subsequently, the divided experiences are randomly morphed by affine transformation to explore the search tree flexibly beyond the range of the original experience path. Owing to these specific local search processes, ERT and ERTC showed significantly better performances in solving difficult planning problems in a cluttered environment than existing planners, even when using only one experience [23].

However, the challenge of retrieval and repair approaches is to increase the solution quality. Because retrieval and repair algorithms focus on finding a valid path in a new environment, the solution quality worsens when a large deformation or additional motion is added to the original experience path. Therefore, increasing the solution quality while maintaining the flexibility of the retrieval and repair processes is a key challenge.

### III. METHOD

### A. Aim and Contribution of This Study

Based on the above research background, the purpose of this study is to address the issue of current motion-planning algorithms in determining optimal motion planning in a cluttered environment using a limited dataset. To achieve the above purpose, the key technical challenge is balancing the flexibility of the motion planner to widely search the C-space beyond the original experience to handle the cluttered environment and the high convergent speed to find a high-quality solution within a short calculation time.

To address this challenge, this study developed a new experience-based optimal motion planning algorithm called Informed ERTC* (IERTC*), which effectively introduces RRT* [3] and Informed RRT* [12] -based optimal planning processes into the retrieval and repair approaches of ERTC [23]. It flexibly explores the search tree in a cluttered environment by retrieving and repairing the experience path as the ERTC algorithm while reducing the path cost by introducing the re-wiring process of the RRT* and informed sampling of the Informed RRT* algorithm.

Although the technical details of the IERTC* will be described in the following sections, the contribution of this study is that it enables the generation of high-quality robot motion in a cluttered environment, where existing optimal motion planning algorithms mostly fail, with a high probability within an acceptable time (e.g., 10 s) using a limited number of experiences (e.g., less than 100). This increases the robot's applicability to environments with greater variability, such as the high-mix, low-volume production line [24], which remains challenging at present because, in these environments, the robot must find a high-quality solution using experiences collected in different situations. The source code of IERTC* is available on the GitHub (https://github.com/takamido/IERTCstar).

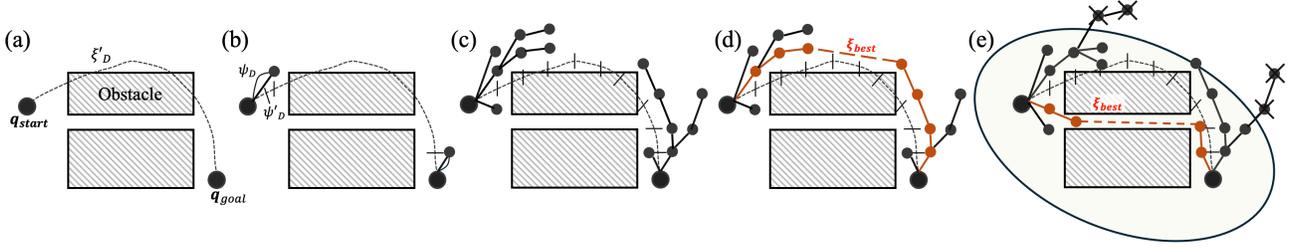

**Fig. 1**. The process of IERTC* consists of the following steps: (a) map the selected experience to the current problem, (b) divide the experience into micro experiences and start the search process while applying the rewiring process at each sampling step, (c) explore the search tree from both directions, (d) find an initial solution, and (e) prune the search trees and restrict the sampling region based on the current solution cost.

*B. Problem definition*

This study targets the motion-planning problem for a single query solved in C-space. The C-space for $n$ degrees of freedom robot is defined as $Q \in \mathbb{R}^n$, and the $Q_{obst} \in Q$ and $Q_{free} = Q \setminus Q_{obst}$ are the part of the C-space occupied by obstacles and collision free part. Specifically, same as the ERTC-based sampling [23], the robot state is defined as $s = \langle q, \alpha \rangle \in \mathbb{R}^{n+1}$ where $q \in Q$ is the particular robot configuration and $\alpha \in [0,1]$ is the phase variable that represents the progress on the execution of a motion plan. The $\alpha$ controls which micro experience within the whole experience path is referenced, and is set as the discrete value corresponding to the number of divisions of the experience path. From a stricter viewpoint, because of this initialization in the search process, the optimality of the RRT*-based process is ensured only when the number of partitions is infinitely large. However, as shown in the results of later experiments, even with this discretization, IERTC* can still significantly reduce the path cost.

Given $S = Q \times \mathbb{R}_{[0,1]}$ is the robot state in C-space, the subset of the collision-free state is defined as

$$S_{free} = \{\langle q, \alpha \rangle \in S \mid q \in Q_{free}\}. \quad (1)$$

Specifically, the path library $A = \{\xi_{D1}, \xi_{D2}, \ldots, \xi_{Dj}\}$ is the set of experience path $\xi_D$ planned in the previous planning problems. This includes a path planned in a different environment from that of the current query. Finally, given path library A, and the valid start $\langle q_{start}, 0 \rangle \in S_{free}$ and goal states $\langle q_{goal}, 1 \rangle \in S_{free}$, the problem solved by the motion planner is defined as to find a collision free path $\xi: \alpha \in [0,1] \to S_{free}$ between the $\xi(0) = q_{start}$ and $\xi(1) = q_{goal}$.

*C. IERTC* algorithm*

Fig. 1 shows a schematic of the IERTC* process, and Algorithms 1-5 show its pseudocode. It is mainly based on the ERTC algorithm [23], and its modified parts are written using red-colored text in the algorithms. In this process, it first selects the most relevant experience from the path library and maps it to the current problem (Fig. 1 (a), Algorithm 1, lines 6 and Algorithm 2, lines 3-7). Similar to a previous study [23], this study selected the experience that had the closest initial and goal configurations to the current query as the reference path. Mapping was performed using the following affine transformation (Algorithm 2, line 15):

---

**Algorithm 1: IERTC*.**

**Input:**
$s_{start}$ and $s_{goal}$ : start and goal states
$\xi_D$: experience path selected from path library

**Output:**
$\xi_{best}$: the final solution path returned by this algorithm

1  # initialization
2  $\xi_{best} \leftarrow \emptyset$; $c_{best} \leftarrow infinite$
3  $\mathcal{T}_{start}.\text{init}(s_{start}); \mathcal{T}_{goal}.\text{init}(s_{goal})$
4
5  # map the experience path onto the current query
6  $\langle \xi'_D, \emptyset \rangle \leftarrow \text{GENERAT\_SEGMENT}(\xi_D, s_{start}, s_{goal})$
7
8  **while** not STOPPING_CONDITION() **do**
9  $\quad s_{init} \leftarrow \mathcal{T}_{start}.\text{select\_node}()$
10
11 $\quad$ # micro experience generation
12 $\quad \langle \psi, s_{targ} \rangle \leftarrow \text{GENERATE\_SEGMENT}(s_{ini}, \emptyset, \xi'_D)$
13
14 $\quad$ **if** $\xi_{best} \neq \emptyset$ **then**
15 $\quad\quad$ **if** REJECT_SAMPLE($s_{targ}, C_{best}$) **then**
16 $\quad\quad\quad$ **continue** # skip remaining processes and back to the top
17
18 $\quad$ # tree extension
19 $\quad$ **if** EXTEND($\mathcal{T}_{start}, \psi, s_{init}, s_{targ}, explore = 1$) $\neq Fail$ **then**
20 $\quad\quad$ **if** OTHER_EXTREME_REACHED($s_{targ}$) **then**
21 $\quad\quad\quad \xi_{best} \leftarrow \text{PATH}(\mathcal{T}_{start})$
22 $\quad\quad\quad c'_{best} \leftarrow \text{CALCULATE\_COST}(\xi_{best})$
23 $\quad\quad\quad$ **if** $c_{best} - c'_{best} < prune\_threshold$ **then**
24 $\quad\quad\quad\quad \text{PRUNE}(\mathcal{T}_{start}, \mathcal{T}_{goal}, c'_{best})$
25
26 $\quad\quad$ # check connectivity to the other tree
27 $\quad\quad s_{near} \leftarrow \mathcal{T}_b.\text{nearest\_neighbour}(s_{targ})$
28 $\quad\quad \langle \psi, s_{targ} \rangle \leftarrow \text{GENERATE\_SEGMENT}(s_{init}, s_{targ}, \xi'_D)$
29 $\quad\quad$ **if** EXTEND($\mathcal{T}_{goal}, \psi, s_{neat}, s_{targ}, explore = 0$) $\neq Fail$ **then**
30 $\quad\quad\quad \xi_{best} \leftarrow \text{PATH}(\mathcal{T}_{start}, \mathcal{T}_{goal})$
31 $\quad\quad\quad c'_{best} \leftarrow \text{CALCULATE\_COST}(\xi_{best})$
32 $\quad\quad\quad$ **if** $c_{best} - c'_{best} < prune\_threshold$ **then**
33 $\quad\quad\quad\quad \text{PRUNE}(\mathcal{T}_{start}, \mathcal{T}_{goal}, c'_{best})$
34
35 $\quad$ SWAP($\mathcal{T}_{start}, \mathcal{T}_{goal}$)
36 **return** $\xi_{best}$

\* red texts represents the changed parts from the original ERTC algorithm

$$\begin{bmatrix} \overline{q} \\ \overline{\alpha} \end{bmatrix} = \begin{bmatrix} \mathbb{I}_{n \times n} & \lambda_{n \times 1} \\ \mathbf{0}_{1 \times n} & |\psi_{D_\alpha}| \end{bmatrix} \begin{bmatrix} \overline{q}_D \\ \rho \end{bmatrix} + \begin{bmatrix} b_{n \times 1} \cdots b_{n \times 1} \\ \alpha_{ini} \cdots \alpha_{end} \end{bmatrix}_{(n+1) \times k}, \quad (2)$$

**Algorithm 2: GENERATE_SEGMENT.**

**Input:**
 $s_{ini}$: required segment configuration-phase start
 $s_{targ}$: required (if any) segment configuration-phase end
 $\xi'_D$: mapped experience path

**Output:**
 $\psi$: generated segment
 $s_{end}$: end configuration of the segment $\psi$

1 $\langle q_{init}, \alpha_{ini} \rangle = s_{ini}$
2
3 **if** not $s_{targ} = \emptyset$ **then** # map the experience
4 $\quad \langle q_{targ}, \alpha_{targ} \rangle = s_{targ}$
5 $\quad \psi_D \leftarrow \xi'_D(\alpha_{init}, \alpha_{targ})$
6 $\quad b \leftarrow q_{init} - \psi_D(\alpha_{init})$
7 $\quad \lambda \leftarrow q_{targ} - (\psi_D(\alpha_{targ}) + b)$
8
9 **else** # generate micro experience
10 $\quad \alpha_{targ} \leftarrow$ SAMPLE_SEGMENT_END($\alpha_{init}$)
11 $\quad \psi_D \leftarrow \xi'_D(\alpha_{init}, \alpha_{targ})$
12 $\quad b \leftarrow q_{init} - \psi_D(\alpha_{init})$
13 $\quad \lambda \leftarrow \mathbb{U}(-\epsilon|\psi_{D_\alpha}|, \epsilon|\psi_{D_\alpha}|)$
14
15 $\psi \leftarrow$ MORPH_SEGMENT($\psi_D, \lambda, b$)
16 $s_{end} \leftarrow \langle \psi(\alpha_{targ}), \alpha_{targ} \rangle$
17 **return** $\langle \psi, s_{end} \rangle$

---

**Algorithm 3: EXTEND**

**Input:**
 $\mathcal{T}$: tree of previously generated micro-experiences
 $\psi$: new generated micro-experience
 $s_{ini}$ and $s_{targ}$: start and end configuration-phase of $\psi$
 *explore*: the bool variable for switching the mode

**Output:**
 outcome of the tree extension attempt

1 **if** IS_VALID($\psi$) **then**
2 $\quad$ **if** *explore* = 1 **then**
3 $\quad\quad$ # check the possibility for the direct connection
4 $\quad\quad$ **if** CHECK_CONNECTIVITY($s_{ini}, s_{targ}$) **then**
5 $\quad\quad\quad$ REPLACE_MOTIOM($\psi$, MOTION($s_{ini}, s_{targ}$))
6
7 $\quad\quad$ # then, check the improvements by rewiring
8 $\quad\quad \langle s_{ini}, \mathcal{T}, \psi \rangle \leftarrow$ REWIRE($s_{ini}, s_{targ}, \mathcal{T}, \psi$)
9
10 $\quad \mathcal{T}$.add_vertex($s_{targ}$)
11 $\quad \mathcal{T}$.add_edge($\psi, s_{ini}, s_{targ}$)
12 $\quad$ **return** *Advance*
13
14 **else**
15 $\quad$ **return** *Fail*

---

**Algorithm 4: REWIRE**

**Input:**
 $s_{init}, s_{targ}$: start and target states
 $\mathcal{T}$: tree of previously generated micro-experiences
 $\psi$: new generated micro-experience

**Output:**
 $s_{targ}$: updated start (parent) state
 $\mathcal{T}$: updated tree
 $\psi$: updated micro-experience

1 $S_{near} =$ Near($s_{targ}, \mathcal{T}, r$)
2 $c_{current} =$ COST_COME($\mathcal{T}, s_{init}$) + COST_GO($s_{init}, s_{targ}$)
3 **forall** $s \in S_{near}$ **do**
4 $\quad c_{new} =$ COST_COME($\mathcal{T}, s$) + COST_GO($s, s_{targ}$)
5 $\quad$ **if** $c_{new} < c_{current}$ **then**
6 $\quad\quad$ **if** CHECK_CONNECTIVITY($s, s_{targ}$) **then**
7 $\quad\quad\quad c_{current} \leftarrow c_{new}$
8 $\quad\quad\quad s_{init} \leftarrow s$
9 $\quad\quad\quad \psi \leftarrow$ MOTION($s_{init}, s_{targ}$)
10
11 **forall** $s \in S_{near}$ **do**
12 $\quad c_{current} =$ COST_COME($\mathcal{T}, s$)
13 $\quad c_{new} =$ COST_COME($\mathcal{T}, s_{targ}$) + COST_GO($s_{targ}, s$)
14 $\quad$ **if** $c_{new} < c_{current}$ **then**
15 $\quad\quad$ REMOVE_MOTIOM($\mathcal{T}$, FIND_MOTIOM($s_{parent}, s$))
16 $\quad\quad$ SET_PARENT($s, s_{targ}$)
17 $\quad\quad \psi_{new} =$ MOTION($s, s_{targ}$)
18 $\quad\quad \mathcal{T}$.add_edge($\psi, s, s_{targ}$)
19
20 **return** $s_{init}, \mathcal{T}, \psi$

---

**Algorithm 5: PRUNE**

**Input:**
 $\mathcal{T}_{start}, \mathcal{T}_{goal}$: start and goal trees
 $c'_{best}$: updated best cost

1 **forall** $\langle s, s_{parent}, \psi \rangle \in \mathcal{T}_{start}, \mathcal{T}_{goal}$ **do**
2 $\quad \hat{c} =$ CALCULATE_HEURISTIC_COST($s, s_{start}, s_{goal}$)
3 $\quad$ **if** $\hat{c} > c'_{best}$ **then**
4 $\quad\quad \mathcal{T} \xleftarrow{-} \langle s, s_{parent}, \psi \rangle$ # remove edge
5 $\quad\quad$ REMOVE_CHILDREN($s, \mathcal{T}$) # remove child's nodes
6 $\quad\quad \mathcal{T} \xleftarrow{-} s$ # remove target state
7

---

where $\lambda_{n \times 1}$ is the shearing coefficient, $b_{n \times 1}$ is a shifting vector, and $\rho = [0, \ldots, 1]_{1 \times k}$ is a local representation of $\bar{\alpha}_D$. By calculating the $\lambda_{n \times 1}$ and $b_{n \times 1}$ which morphs the experience path $\xi_D$ to match the initial and goal configuration of the current, the experience path $\xi'_D$ is obtained (Fig.1 (a)).

Subsequently, the search trees were explored from both the start and goal configurations in the C-space (Algorithm 1, line 8, Fig. 2 (b)–(c)). When exploring the search tree, the divided micro-experiences are randomly morphed to flexibly search the C-space while avoiding obstacles. This morphing process enables the planner to generalize and adjust the experience path to an environment different from the original one; however, it sometimes generates an unnecessary curved edge, even in cases where no obstacles exist between the initial and target configurations. Hence, to reduce the path cost, the possibility of a direct (straight-line) connection was checked for every sampling step (Algorithm 3, lines 3-5, Fig. 1 (b)). Furthermore, a rewiring process is introduced into the original ERTC algorithm to asymptotically reduce the solution cost (Algorithms 3, 8, and 4). Similar to the original RRT* algorithm [3], it checks two possibilities for improvements by re-wiring among neighboring nodes of the new sampling point: (1) reduction of the reaching cost to the sampling point by changing its parent nodes (Algorithm 4, lines 3-10), (2) reduction of the reaching cost to the existing nodes by changing the parent node to the new sampling point (Algorithm 4, lines 11-18).

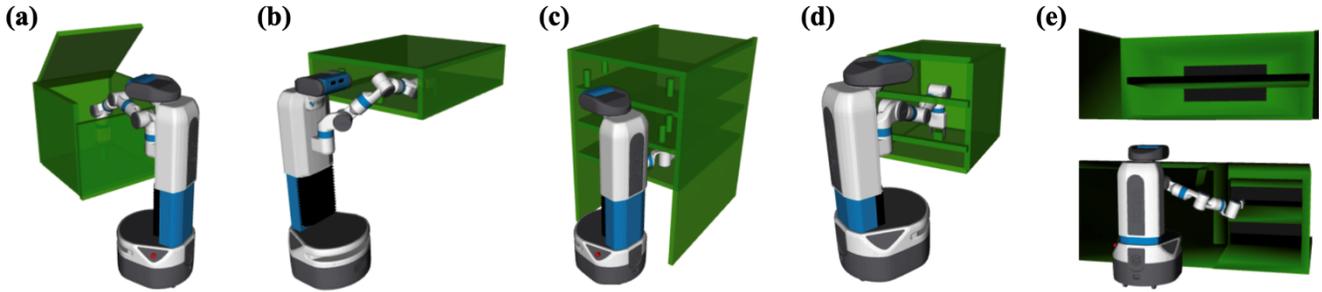

**Fig. 2**. The motion planning environment and queries generated by the Motionbenchmaker [25] and used in the experiment in this study. (a) box, (b) bookshelf-small, (c) bookshelf-tall, (d) cage, (e) kitchen. The robot in each scene shows the goal configuration of the generated query.

Here, the point of the above approach is to use different strategies depending on the complexity of the local environment; that is, it describes a more complex curved trajectory if the obstacles are densely arranged and describes a simpler straight line if the local environment is sparse.

After finding a solution that has a lower cost than the current one, the pruning process is started to remove the redundant nodes in the search trees (Algorithm 1, lines 23-24 and 32-33, Algorithm 5), and the sampling region is restricted, which has the potential to improve the current cost (Fig1. (e) Algorithm 1, lines 14-16). By iterating these processes, the IERTC* algorithm can asymptotically improve the solution quality, similar to RRT*-based algorithms.

## IV. BENCHMARK EXPERIMENT

A benchmark experiment was conducted to verify the effectiveness of the proposed method. In this study, the Motionbenchmaker developed by [25] was used for testing because it provides various difficult planning problems in a cluttered environment, such as picking up the object in the bookshelf, as shown in Fig. 2, and can add variations to planning queries, such as changing the positions and poses of the objects in the environment. Using Motionbenchmaker, one thousand planning problems were generated for each of the five scenarios shown in Fig. 2 ((a) box, (b) tall bookshelf, (c) small bookshelf, (d) cage, and (e) kitchen). Among these queries, the bookshelf-tall, bookshelf-small, and cage scenarios are especially difficult environments where robots need to move their arms to pass through a very narrow space. To add variations to the generated queries, the start and goal configurations of the robot and the positions and poses of the objects in the environment were randomly changed during the generation process. The range of variations was set as the default values of the Motionbenchmaker for each scenario. The feasibility of the generated problems was checked by solving the bidirectional kinematic motion planning by interior-exterior cell exploration (BKPICE) [26] with a 60 s calculation, and its solution costs were recorded as the criterion for the path cost evaluation of the optimal motion planning. A model of the fetch robot (Fetch Robotics, Inc.) with eight degrees of freedom was selected for the target robot, considering the high dimensionality of the C-space. The reason for using BKPICE for data generation was that it demonstrated the highest performance in Motionbenchmaker with the Fetch robot in a previous study [27].

IERTC* was implemented on the ROS Noetic with OMPL Open Motion Planning library [28]) platform. To generate the path library for the IERTC*, 100 planning problems were randomly generated for each scenario, and the solution paths solved by BKPICE were saved as experiences. To verify the ability to generalize a limited number of experiences, a two-type algorithm was implemented: IERTC*-100, which uses 100 experiences, and IERTC*-1, which only uses one experience randomly selected from 100 experiences. Because most previous studies used more than 1000 datasets (e.g., [7, 19]), it is worthwhile that the IERTC* can facilitate the optimal motion planning process by using those limited datasets.

Additionally, to make comparisons with existing optimal motion planners, RRT* [3], BIT* [13], AIT* [14], and GMR-RRT* [16] were implemented. BIT* is the optimal motion planner, which is an improvement from Informed-RRT* by adding several improvements, such as introducing a batch process. AIT* is a state-of-the-art planning algorithm developed from BIT* by introducing a more adaptive heuristic cost function. The GMR-RRT* is a state-of-the-art model-based planning algorithm that performs biased sampling using the GMR model. The GMR-RRT* models were built using 100 experiences from the path library. To fairly compare the performance of each planner, the hyperparameters for each planner, such as the radius to define the neighborhood for the RRT*-based sampling, were set as the default values of the OMPL and the original paper. The planning time was set to 10 s. The simulation ran on an Intel i5 Linux machine with 2.0 GHz cores and 16.0 GB RAM. The success rate of the 10 s planning and the path length were used as evaluation indexes. The path length was evaluated by calculating the reduction rate relative to that generated by the BKPICE during the dataset creation process.

## V. RESULTS AND DISCUSSIONS

Figs. 3 and 4 show the results of the benchmark tests. As shown in Fig. 3, IERTC* exhibited the highest success rate in all scenarios. In comparison with IERTC*-100 and AIT*, which showed the best success rates among the existing planners, improvements of 9.5 %, 75.9 %, 74.2 %, 59.5 %, and 27.5% were observed in each scenario. Furthermore, the results of IERTC*-1, which used only one experience to solve one thousand problems, also showed significant improvements in the success rate (9.5%, 72.5%, 69.0%, 43.3%, and 24.7%, respectively). These results demonstrate the effectiveness of the proposed algorithm for solving difficult motion-planning problems in a cluttered environment using a limited dataset.

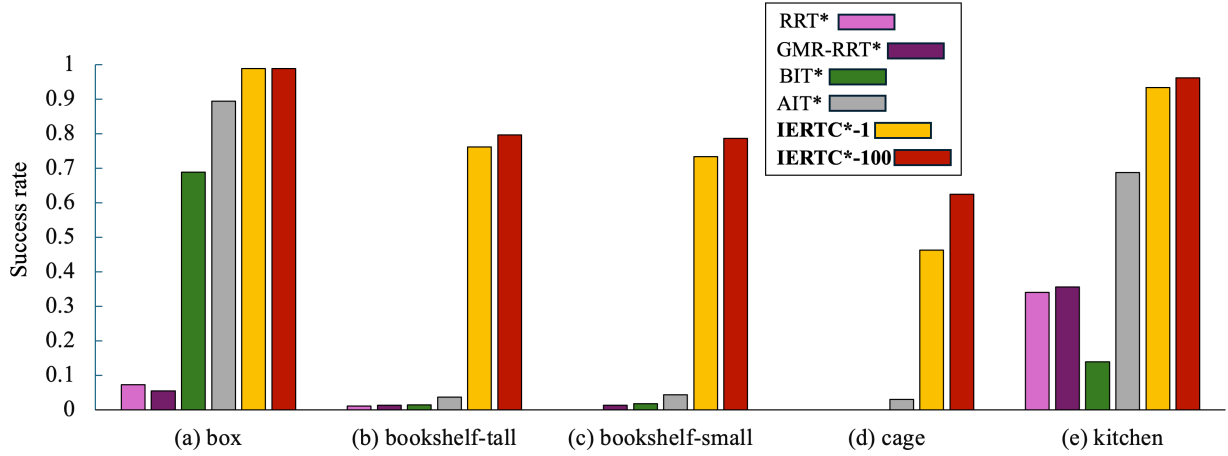

**Fig. 3**. The success rate of the motion planners in each five-scenario shown in Fig. 2.

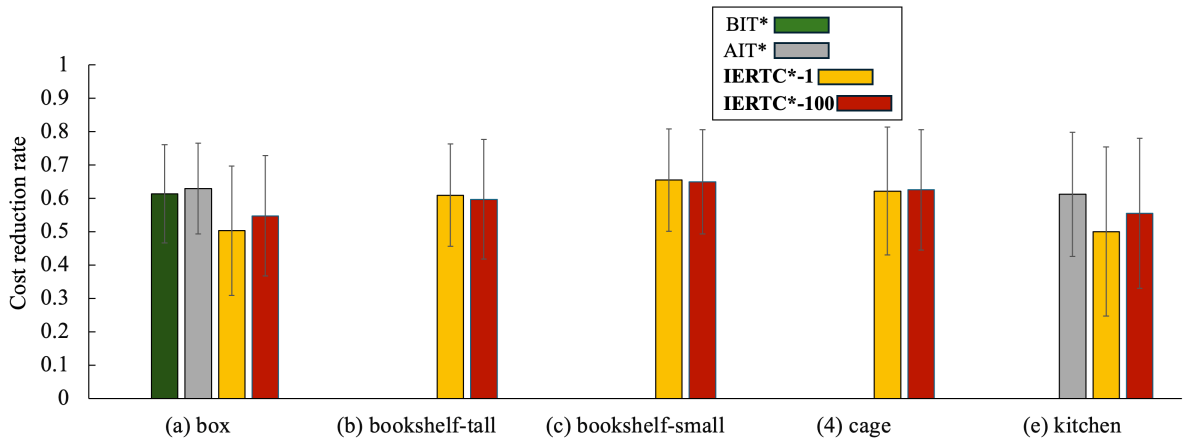

**Fig. 4**. The relative cost reduction rate to the BIKPICE of each planner. Only the data of the planner which revealed more than 50 % success rate (Fig. 3) was showed in this figure.

More specifically, the effectiveness of the proposed methods is greater in the difficult planning scenarios of bookshelf-tall, bookshelf-small, and cage (Figs. 2. (b)–(d)) than in the simpler scenarios (Figs. 2. (a) and (e)). In these difficult scenarios, whereas other existing planners mostly failed to plan, the proposed methods showed 50-80% success rates. This suggests that the use of past experiences is more effective in a complex environment where existing planners have difficulty planning motions. As indicated in previous studies [23, 29], ERTC-based local search strategies that flexibly and intensively explore the area around existing nodes are more effective in cluttered environments with many surrounding obstacles.

Regarding the solution cost as a result of the optimal motion planning attempts, ERTC*-100 showed 54.7%, 59.2%, 65.0%, 62.5%, and 55.5% reductions in each scenario from the path length of BKPICE for the same problems (Fig. 4). Therefore, IERTC* also significantly improves the solution quality in all scenarios, even when it can only use a single experience (IERTC*-1). However, comparisons with the existing optimal motion planners showed 6-8% worser optimization performance than AIT* or BIT* in the box and kitchen scenarios (Figs. 4 (a) and (e)).

To explore this point in more detail, we analyzed the convergence speed of the path cost in the box scenario (Fig. 5). As shown in Fig. 5, although IERTC* initially shows a sharp decline in optimization cost, it is overtaken by other methods after approximately 3 s. There are several possible reasons for this observation. First, because the algorithm running inside IERTC* is relatively complex compared to other motion-planning algorithms such as AIT*, this may affect the latter part of the planning. From our observations, while the AIT* iterates the sampling loop for several tens of thousands during the 10 sec planning, the IERTC* can only perform for several thousands. Therefore, it is necessary to add further improvements to increase the sampling efficiency, such as introducing laundry collision checking [10]. Furthermore, as described in the Methods section, because IERTC* uses the discrete search strategy with divided micropaths, it is possible that this affects the quality of the final cost.

Additionally, from the comparisons between ERTC*-100 and ERTC*-1, the effect of the size of the dataset increases for a difficult scenario such as a cage (Fig. 3 (d)). This suggests that the required dataset size changes depending on the environmental complexity. Therefore, if the problem becomes more difficult than the one in this study, a larger dataset may be required.

Finally, this study had some limitations. First, as previously described, the sampling efficiency of the proposed algorithm should be improved to increase the quality of the final solution. Additionally, because this study adopted a simple similarity function to select experiences from the path library, a more sophisticated function should be developed and incorporated, as in [30]. Finally, from a practical perspective, it is necessary to develop an entire framework for planning the motion, saving it as an experience, and removing the unnecessary experience based on it in an online manner, as in some studies [21, 31].

## VI. CONCLUSION

In conclusion, the proposed IERTC*algorithm significantly improved the success rate of motion planning (49.3% improvement on average for five scenarios) while also significantly improving the solution quality (56.3% reduction on average to the cost of BKPICE) by using a limited dataset. Furthermore, it revealed a high ability to generalize one experience to solve one thousands planning problems. Although the proposed method has some limitations, such as algorithmic complexity, this study addresses a critical issue in current motion planning — how to find a high-quality solution path within a short calculation time for difficult planning problems using a limited dataset. This enhances the applicability of robotic systems in environments with greater variability, which remains a challenging task.

## VII. ACKNOWLEDGMENT

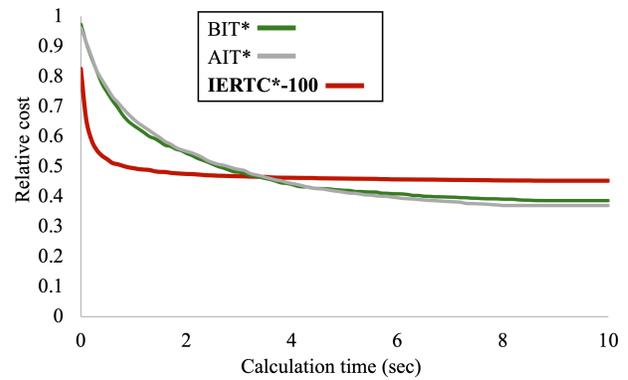

**Fig. 5**. The relationship between calculation time and solution quality is shown. The y-axis represents the relative cost compared to the path length of BKPICE. The average values among successful cases were calculated and plotted in this figure.